\documentclass[runningheads]{llncs}
\usepackage{graphicx}
\usepackage{amsmath} 
\usepackage{booktabs}
\usepackage{amssymb}

\begin{document}

\title{Enhancing Texture Generation with High-Fidelity Using Advanced Texture Priors} 

\titlerunning{HD-TEXTure}




\author{Kuo Xu\inst{1,2} \and
Maoyu Wang\inst{1,}\thanks{Equal contribution.} \and
Muyu Wang\inst{1,3} \and
Lincong Feng\inst{1,4} \and \\
Tianhui Zhang\inst{1} \and 
Xiaoli Liu\inst{5,}\thanks{Corresponding author.} }

\authorrunning{Xu,~Kuo et al.}

\institute{MetaApp AI Research, Beijing, China \and
Zhengzhou University, Zhenzhou, China
\and
Beijing Institute of Technology, Beijing, China
\and
Beijing University of Technology, Beijing, China
\and
Northeastern University, Shenyang, China
}

\maketitle

\begin{abstract}
 The recent advancements in 2D generation technology have sparked a widespread discussion on using 2D priors for 3D shape and texture content generation. However, these methods often overlook the subsequent user operations, such as texture aliasing and blurring that occur when the user acquires the 3D model and simplifies its structure. Traditional graphics methods partially alleviate this issue, but recent texture synthesis technologies fail to ensure consistency with the original model's appearance and cannot achieve high-fidelity restoration. Moreover, background noise frequently arises in high-resolution texture synthesis, limiting the practical application of these generation technologies.In this work, we propose a high-resolution and high-fidelity texture restoration technique that uses the rough texture as the initial input to enhance the consistency between the synthetic texture and the initial texture, thereby overcoming the issues of aliasing and blurring caused by the user's structure simplification operations. Additionally, we introduce a background noise smoothing technique based on a self-supervised scheme to address the noise problem in current high-resolution texture synthesis schemes. Our approach enables high-resolution texture synthesis, paving the way for high-definition and high-detail texture synthesis technology. Experiments demonstrate that our scheme outperforms currently known schemes in high-fidelity texture recovery under high-resolution conditions. Detailed code and supplementary visual effects can be found at: \url{https://hd-texture.github.io/}
\end{abstract}

\section{Introduction}
\label{sec:intro}

Recently, generative artificial intelligence technology has garnered widespread attention in the industry. While these technologies have shown impressive results in 2D generation, the field of 3D generation still requires further development. 3D assets are crucial for games, movies, and VR/AR, but they often need to be manually crafted by professional graphic designers. With the rise of the metaverse concept, traditional handcrafting will not suffice to meet the demand for massive 3D assets in emerging virtual technologies. To address this urgent need, automated 3D generation techniques have been extensively researched \cite{chan2022efficient,chen2022gdna,dong2023ag3d,ma2020learning,park2019deepsdf,pavlakos2019expressive,wu2016learning}. Notably, text or image-guided 3D asset generation based on neural radiance fields and generative diffusion models \cite{huang2024dreamwaltz,jiang2023avatarcraft,lin2023magic3d,poole2022dreamfusion} has recently become mainstream. To use these assets in a real rendering engine (Blender or Unity), we need to transform the implicit representation into an explicit mesh and the corresponding texture map. The classical extraction algorithm is the Marching Cubes algorithm, and high surface is often adopted for more accurate mesh geometry extraction.

\begin{figure}[t]
  \centering
  \includegraphics[width=\linewidth]{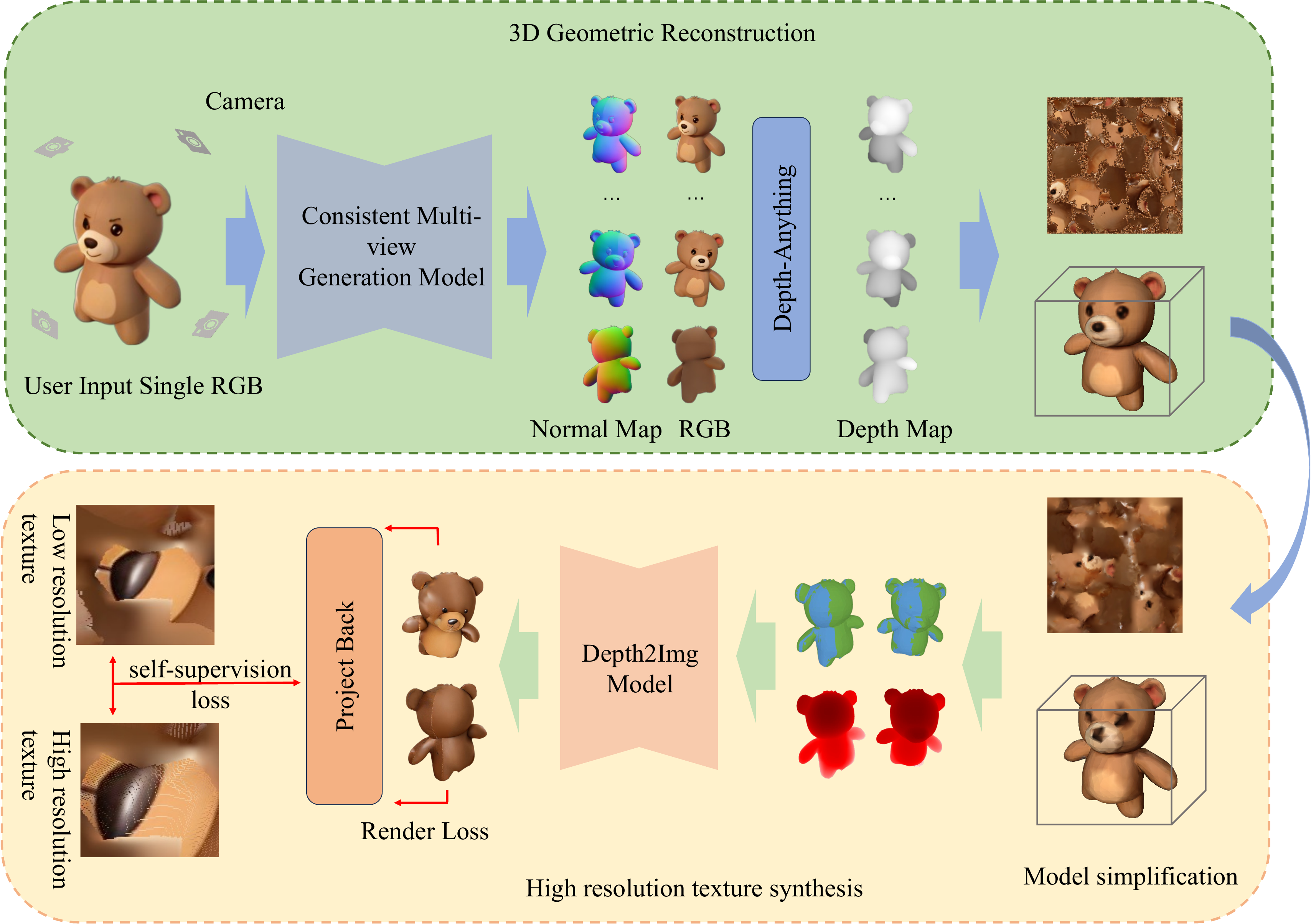}
  \caption{The overall framework of our method comprises two stages. In the first stage of geometric restoration, we generate RGB images with multi-view consistency and corresponding normal maps for a single input image, incorporating depth supervision using the Depth anything model. This stage primarily serves to provide users with textured meshes for structural simplification, is not our focus. In the second stage, we begin by rendering the rough texture post-user operation as the initial input. We then generate an RGB image using the Depth2Img model and project it onto the texture map for gradient optimization. During this projection, we produce both low-resolution and high-resolution textures for self-supervision, effectively eliminating noise and "point gaps" in the high-resolution texture map.}
  \label{fig:pipline}
\end{figure}
After obtaining a high-face mesh and the corresponding texture map, users can perform operations in the real rendering engine, such as reducing faces and stretching as needed. Texture distortion, aliasing, and blurring are common problems associated with user manipulation. Traditional graphics techniques \cite{garland1997surface,zhang2022seamless,coll2010accurate,xian2020mesh,bahirat2017boundary} can alleviate these issues, but cannot completely overcome them. Recent technology \cite{richardson2023texture,chen2023text2tex,liu2023text,youwang2023paint} for generating textures for known meshes based on text input is rapidly developing, but it is often constrained by the input text description and lacks user control over existing textures. Inspired by neural network generation methods, we use the user's existing rough texture as the initial rendering input, combined with the corresponding text description, to constrain the generated content. This approach enhances both the consistency and controllability of the generated content.

We also note that recent texture generation schemes for white models tend to produce severe noise in high-resolution conditions, limiting practical use in real rendering engines. We analyze the causes of this noise and propose a self-supervised solution based on high and low resolution. Experimental comparisons with state-of-the-art open-source schemes demonstrate the effectiveness of our module.

In summary, our main contributions are :
\begin{enumerate}
    \item[-] We propose a neural network generation scheme using initial input to overcome the aliasing and blurring problems caused by mesh reduction, and introduce a new solution for traditional graphics techniques.
    \item[-] We propose a scheme based on self-supervision of high and low resolution, which overcomes the noise problem of current texture generation schemes in generating high-resolution textures, and provides conditions for high-quality, high-detail texture generation.
    \item[-] In general, we propose a high fidelity and high quality texture generation scheme, which solves the problem of aliasing and blurring caused by mesh operation, and solves the noise problem when generating high-resolution textures. Our method similarly shows excellent performance when generating textures for white molds without initialized textures.
\end{enumerate}

\section{Related Work}

\subsection{3D Reconstruction Methods}
Recent advancements in 2D generative models, such as DALL-E \cite{ramesh2021zero,ramesh2022hierarchical} and Stable Diffusion \cite{rombach2022high}, along with visual language models like CLIP \cite{radford2021learning}, have achieved impressive visuals using massive images from the Internet. However, the lack of public 3D data hinders direct training of large models for 3D generation to produce effects comparable to 2D visual fields. Nonetheless, the use of 2D priors to assist 3D generation tasks has proven effective. Approaches like DreamFiled \cite{jain2022zero} and DreamFusion \cite{poole2022dreamfusion} optimize a 3D representation (like NeRF) and generate 2D images from different perspectives using microrendering. These images are then used with 2D diffusion models \cite{feng2023metadreamer,lin2023magic3d,seo2023let,melas2023realfusion,deng2023nerdi,wang2023score,xu2023neurallift,metzer2023latent,zhou2023sparsefusion,raj2023dreambooth3d} or the CLIP model \cite{hong2022avatarclip,jain2022zero,michel2022text2mesh,lee2022understanding,canfes2023text,khalid2022text,aneja2023clipface,jetchev2021clipmatrix,xu2023dream3d,liu2022iss} to calculate loss functions and guide the optimization of 3D shapes and appearances. However, this process is often slow and prone to multi-face problems due to a lack of clear 3D understanding.

There is also work on generating multi-view images for NeRF with view consistency, which often requires fine-tuning the Stable Diffusion model. Zero123 \cite{liu2023zero} controls the diffusion model to generate content from the corresponding view using the relative camera position as an additional input. Watson et al. \cite{watson2022novel} first applied the diffusion model on the ShapeNet dataset \cite{sitzmann2019scene} for new view synthesis. GeNVS \cite{chan2023genvs} enhances view consistency by re-projecting underlying features during the denoising process. However, these methods are often limited by the capacity boundary of the training set and cannot be generalized to different image inputs. To overcome dataset limitations, Viewset Diffusion SyncDream \cite{szymanowicz2023viewset} and MVDream \cite{shi2023mvdream} propose a method to adjust the attention layer to produce coherent multi-view color images. Wonder3D \cite{long2023wonder3d} adopted this attention layer adjustment scheme, but also noted that 3D reconstruction schemes based on color images often have texture ambiguity and lack geometric details. Therefore, Wonder3D proposed using view-consistent normal lines for geometric restoration, achieving alignment between normals and color images by adjusting cross-domain attention and enhancing geometric reconstruction details. We made minor adjustments to Wonder3D's approach to increase its robustness, but please note that this is not our primary focus.

\subsection{Texture Synthesis Methods}
In addition to text or image-guided 3D geometry generation, generating reasonable textures for existing meshes is also crucial. Early research \cite{ashikhmin2001synthesizing,efros1999texture,kopf2007solid,kwatra2005texture} focused on creating 2D and 3D tileable texture maps from samples and applying them to objects. Some studies \cite{lu2007context,mertens2006texture,wong1997geometry} explored the correlation between surface structure and texture details, enabling geometric perception synthesis and the transfer of texture details. However, both rule-based approaches and machine learning-based models have limited capacity.

Recently, texture synthesis based on generative models has become mainstream. Some works \cite{mohammad2022clip,sanghi2022clip} utilize the 2D prior of generative models, extract gradients from the CLIP model, use contrast learning to compare text with rendered images for training, and constantly update the rendering view of the 3D object through gradient descent to make the texture map's rendering effect conform to the given text. Texture \cite{richardson2023texture} and Text2Tex \cite{chen2023text2tex} use the depth conditional diffusion model for progressive repair in texture synthesis. In this approach, depth is first rendered from a certain perspective, and both the rendering depth and description text are used as diffusion conditions to generate 2D views. The generated images from this perspective are then reflected into the texture map to fill part of the visible content. By continuously updating the rendering perspective, the entire texture map can be incrementally updated. This method can quickly present clear textures, but it is affected by texture seams and inconsistent views.

TexFusion \cite{cao2023texfusion} attempts to perform a round of projections and fixes in each de-noising timestep to resolve inconsistencies. SyncMVD \cite{liu2023text} proposes a method for sharing potential information from all perspectives, fusing information from all views, reaching consensus on content structure and color distribution, and reducing seams and inconsistencies. These methods bring new ideas to texture synthesis, but they are limited by resolution and generate uncontrollable background noise when generating content at high resolution (1K, 2K), which limits their practical use. We propose a known self-supervised method to overcome this problem.

\subsection{Traditional mesh and texture simplification algorithms.}
Generative model-based 3D generation and texture synthesis methods enable the free creation of 3D content. Users can freely build new content by combining these two methods, but when performing operations on the generated content (such as surface reduction), texture jagging and blurring often occur. Texture synthesis methods \cite{richardson2023texture,chen2023text2tex,cao2023texfusion,liu2023text} can regenerate new textures but lack control over the original texture map and struggle to maintain the original appearance.

In the fields of computer graphics and 3D modeling, mesh simplification technology is crucial for real-time graphics rendering, virtual reality, augmented reality, online 3D visual display, and game development, as these applications often need to quickly process complex 3D scenes to ensure a smooth user experience \cite{hoppe1997view,garland1997surface}. Therefore, maintaining the original appearance while simplifying the complexity of 3D models has become a research focus. Hoppe \cite{hoppe1997view} introduced Progressive Meshes technology for multi-resolution model representation through continuous refinement. Garland and Heckbert \cite{garland1997surface} presented a mesh simplification method based on Quadric Error Metrics (QEM), effectively preserving model geometry by minimizing shape errors caused by vertex merging. Sander et al. \cite{sander2001texture} explored texture mapping progressive meshes, aiming to preserve texture detail during simplification. Zhang et al. \cite{zhang2022seamless} focused on seamless simplification of multi-graph texture meshes, maintaining texture continuity and overall appearance. Coll et al. \cite{coll2010accurate} proposed a method to accurately simplify models with multi-graph textures while preserving texture details and color information. Xian et al. \cite{xian2020mesh} ensured visual consistency between the simplified model and the original model through appearance-driven optimization. Bahirat et al. \cite{bahirat2017boundary} proposed a new mesh simplification algorithm emphasizing model boundary and texture feature preservation. Despite significant progress, reducing computational complexity while preserving model visual effects remains an open and challenging problem.

Our method integrates the generative model-based texture generation scheme into the model simplification field while maintaining appearance. The rough rendered image after model simplification is used as the initial input to control the consistency and fidelity of the generated content.

\section{Method}
We introduce a high-fidelity 3D recovery solution that separates geometric reconstruction from texture restoration, enabling users to manipulate geometry and achieve high-quality textures post-manipulation. The approach comprises two phases: In the first phase, we recover geometry from a single image, allowing user-driven geometric edits such as mesh simplification and stretching, as outlined in \ref{sec:3D Geometric Reconstruction}. In the second phase, we conduct high-fidelity texture restoration on the modified geometry, detailed in \ref{sec:High-Fidelity Texture Generation}.

\subsection{3D Geometric Reconstruction}
\label{sec:3D Geometric Reconstruction}
The advent of Stable Diffusion and Neural Radiance Fields (NeRF\cite{mildenhall2021nerf}) has made 3D recovery from a single image viable. However, optimizing each viewpoint separately often results in the multi-face Janus issue. A viable solution is using the Stable Diffusion model to generate consistent multi-view images. With these images, Signed Distance Fields (SDF) can be employed for precise multi-view reconstruction.

\subsubsection{Consistent Multi-view Generation.}
To tackle multi-view consistency in image generation with 2D Diffusion models, Dreamfusion introduces additional perspective details. The One-2-3-45++ \cite{liu2023one} model enhances the Diffusion model to simultaneously produce six unique perspectives of an object. Similarly, MVDream \cite{shi2023mvdream} presents a novel approach by modifying the Diffusion Unet's attention module for consistent multi-view image synthesis. We adopt this strategy, specifically altering the 2D attention module to link multiple views. The original 2D attention mechanism used key, query, and value tuples to capture contextual pixel information within an image. We extend this mechanism to multiple views to facilitate inter-view information exchange while maintaining consistency, implementing this on a batch level. Wonder3D \cite{long2023wonder3d} takes this idea further by introducing a cross-domain attention scheme, allowing for the simultaneous generation of coherent multi-view RGB images and corresponding normal maps. We enhance Wonder3D to increase its robustness.

\subsubsection{3D Generation in Multiple View conditions.}
After obtaining consistent multi-view images, we employ Signed Distance Fields (SDF) for reconstructing both the geometry and appearance of the scene. SDF is particularly effective in accurately replicating appearances with indentations. Following the methodology in NeUS, we use two separate networks to predict the SDF values, $\hat s$, and color values, $\hat{\mathbf{c}}$, for points in 3D space.

\begin{equation}\hat s=f_\theta(\gamma(\mathbf{x}),\text{interp}(\mathbf{x},\Phi_\theta)) \qquad \hat{\mathbf{c}}=\mathbf{c}_\theta(\mathbf{x},\mathbf{v},\hat{\mathbf{n}},\hat{\mathbf{z}})
\end{equation}

The feature decoding network, represented by $f_\theta$, is a multilayer perceptron (MLP). The feature vectors for the SDF are stored in a feature grid $ \Phi_\theta \in \mathbb{R}^3 $. When querying the SDF value $\hat s$ of a point $\mathbf{x}$, we interpolate within the feature grid to obtain the feature vector as the input condition. $\gamma(\cdot)$ denotes positional encoding. In the color formulation, $\mathbf{v}$ represents the view direction, and $\hat{\mathbf{n}}$ denotes the unit normal, which is derived by calculating the gradient of the SDF. $\hat{\mathbf{z}}$ represents the feature vector obtained through interpolation. $\mathbf{c}_\theta$ is a parameterized MLP network. To facilitate inverse rendering using differentiable volumetric rendering, we first convert the SDF values into density values based on \cite{yariv2021volume}.

\begin{equation}
\sigma_\beta(s)=\begin{cases}\frac{1}{2\beta}\exp\left(\frac{s}{\beta}\right)&\text{if} \quad s\leq0\\\frac{1}{\beta}\left(1-\frac{1}{2}\exp\left(-\frac{s}{\beta}\right)\right)&\text{if}\quad s>0\end{cases},
\end{equation}
where $\beta$ is a learnable parameter. Then, we compute the final pixel colors using the method of volumetric rendering:

\begin{equation}
\hat{C}(\mathbf{r})=\sum_{i=1}^MT_\mathbf{r}^i\alpha_\mathbf{r}^i\hat{\mathbf{c}}_\mathbf{r}^i\quad T_\mathbf{r}^i=\prod_{j=1}^{i-1}\left(1-\alpha_\mathbf{r}^j\right)\quad\alpha_\mathbf{r}^i=1-\exp\left(-\sigma_\mathbf{r}^i\delta_\mathbf{r}^i\right)\quad
\end{equation}

We modify the volumetric rendering formula to render the depth and normals at the points of intersection with the surface.
\begin{equation}
\hat{D}(\mathbf{r})=\sum_{i=1}^MT_\mathbf{r}^i\alpha_\mathbf{r}^it_\mathbf{r}^i\quad\hat{N}(\mathbf{r})=\sum_{i=1}^MT_\mathbf{r}^i\alpha_\mathbf{r}^i\hat{\mathbf{n}}_\mathbf{r}^i
\end{equation}
We utilize a pre-trained Depth-anything \cite{yang2024depth} model to predict depth maps $D$ of our generated consistent multi-view images within the same batch, serving as pseudo ground-truth to supervise the rendering results. We observed that Wonder3D generates both RGB and normal maps, and reconstructs geometry from the normal maps. Relying solely on normals is susceptible to noise. We anticipate that globally perceived depth and normals, which contain local details, can complement each other.
\subsubsection{Loss function}
Similar to most approaches, we first compute an RGB reconstruction loss to constrain the final predicted color. we note that the depth predicted by Depth-anything needs to be aligned with our rendered depth for consistency to work. Specifically, we process the render depth using a set of scale alignment parameters $(w, q)$, and then calculate the depth consistency loss:
\begin{equation}
\mathcal{L}_{\mathrm{rgb}}=\sum_{\mathbf{r}\in\mathcal{R}}\|\hat{C}(\mathbf{r})-C(\mathbf{r})\|^2 
\qquad 
\mathcal{L}_{\mathrm{depth}}=\sum_{\mathbf{r}\in\mathcal{R}}\left\|(w\hat{D}(\mathbf{r})+q)-{D}(\mathbf{r})\right\|^2
\end{equation}
Considering depth as a global geometric constraint, we also introduce a normal consistency loss to enhance the reconstruction effect of local details like Wonder3D:
\begin{equation}\mathcal{L}_{\mathrm{normal}}=\sum_{\mathbf{r}\in\mathcal{R}}\|\hat{N}(\mathbf{r})-\bar{N}(\mathbf{r})\|_1+\|1-\hat{N}(\mathbf{r})^\top\bar{N}(\mathbf{r})\|_1.\end{equation}
The final reconstruction loss is expressed as:
\begin{equation}\mathcal{L}=\mathcal{L}_{\text{rgb}} + \lambda _1\mathcal{L}_{\text{depth}} + \lambda _2 \mathcal{L}_{\text{normal}} .\end{equation}

\begin{figure}[tb]
  \centering
  \includegraphics[width=0.8\linewidth]{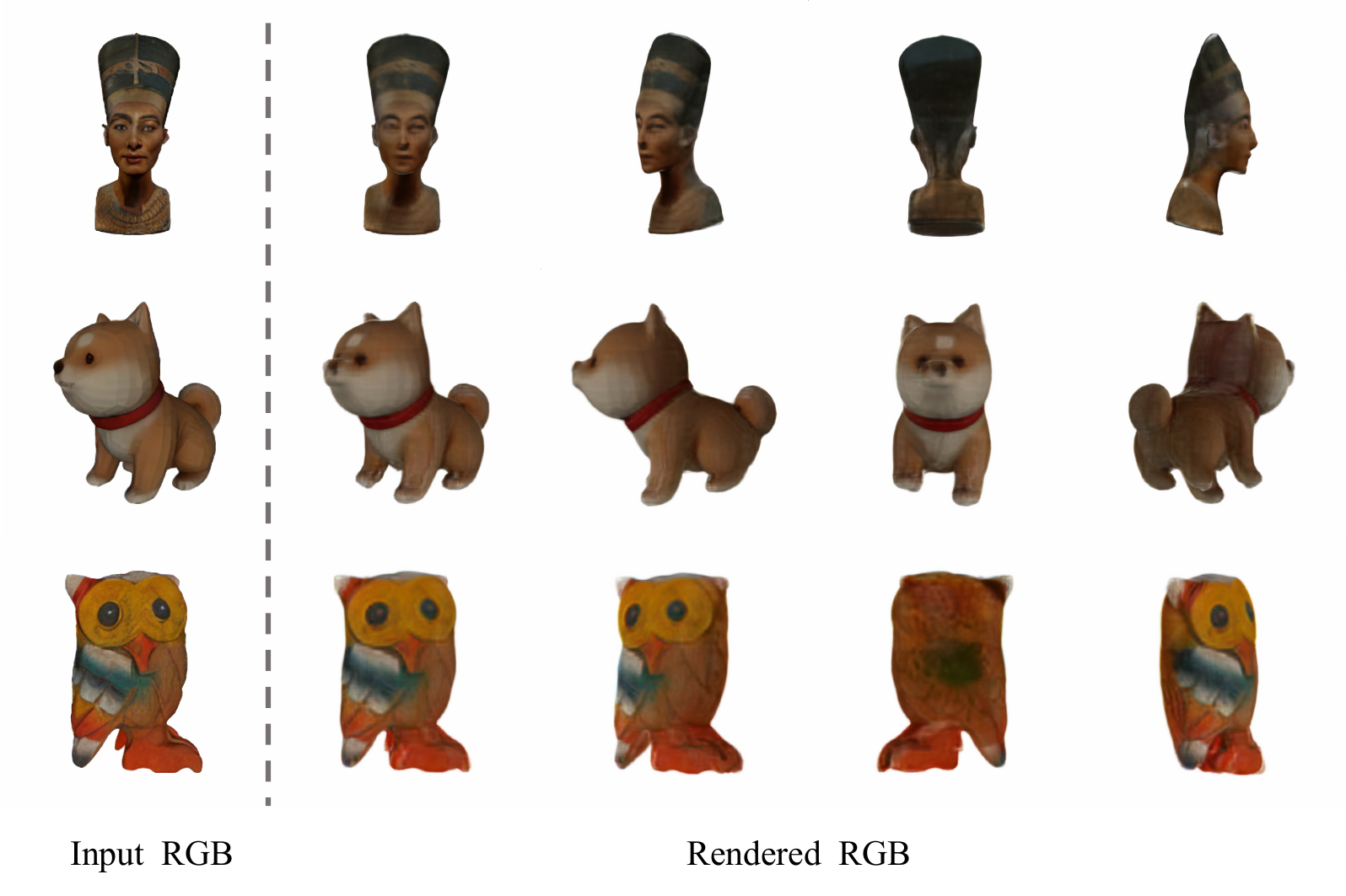}
  \caption{We show the NeRF rendering outcomes from a single-view input; the final rendering quality hinges on the efficacy of generating new views with view consistency.}
  \label{fig:result1}
\end{figure}

\subsection{High-Fidelity Texture Generation}
\label{sec:High-Fidelity Texture Generation}
Upon completing the single-view geometric reconstruction in the initial phase, users can engage in mesh modification tasks such as decimation and stretching. Traditional decimation techniques somewhat mitigate the issue of jagged edges in mesh simplification but fail to entirely eliminate it. The advent of Stable introduces an innovative approach to addressing these challenges in mesh structure simplification. We employ the Depth2Image model to create images consistent across multiple views for the mesh, which are then projected onto the texture map. This texture map undergoes optimization via gradient descent, facilitating high-fidelity texture recovery for meshes that have undergone surface reduction or other modifications leading to diminished surface texture quality.

\subsubsection{Creating Rendering RGB through Depth.}
In the initial stage of geometry and texture recovery, the redesigned batch-self-attention scheme in the Diffusion model simultaneously generates images from multiple views, significantly reducing inconsistency. However, during the high-fidelity texture recovery phase, the Depth2Image model is employed to generate high-quality renderings for each view, yet it cannot fully address the Janus problem in every independent generation. (To ensure our model's versatility, we assume the mesh can originate from any source, thus disregarding multi-view consistent images generated in the first stage.)

Stable Diffusion is known for its de-noising from noise capability. Pseudo-gaussian noise is created by adding noise to input content during training, and the final content is obtained through multi-step de-noising from this noise. Leveraging this feature, for high-fidelity texture recovery in each perspective, we first render the mesh with user-provided fuzzy texture input, obtain the initial input $I_i$, and encode it to obtain $z_i$, retaining input features. By preserving these features, our generated content closely matches the user-provided mesh input. To mitigate inconsistencies further, we input depth images from two symmetrical viewpoints into the Depth2Image model, using this approach as a condition to obtain diffusion images under these perspectives.

\begin{equation}
(\hat{I}_1,\hat{I}_2) = \mathrm{DDPM} (\mathcal{N}((z_1, z_2))|D )  
\end{equation}

Specifically, our image generation process involves three steps: First, we use the processed texture as the initial texture map for rendering, obtaining rendered images $I_1$ and $I_2$ from viewpoints $V_1$ and $V_2$. We then add noise to $I_1$ and $I_2$ for $T$ times, serving as the de-noising input for the Depth2Image Diffusion model. Secondly, we render depth maps $D_1$ and $D_2$ and splice them together as the final input condition $D$. Finally, we calculate the update and maintain regions through the Mesh surface normal, generating content in the update region while preserving the maintain region.

\subsubsection{Divide the Keep area and Update area.}
Upon obtaining the mesh in the initial stage, we partition the mesh area considering that visual effects vary across angles. Specifically, the image $I_i$ rendered at angle $V_i$ is divided into two zones: the keep area and the update area. The keep area maintains superior performance in alternative perspectives, whereas the update area excels in the current view. This division is based on surface normals, designating regions with an angle greater than $\pi/5$ between the normal and the viewpoint as the update area, while others form the keep area. In the keep area, the Depth2Image diffusion model applies a single noise addition and denoising operation to preserve the current view. Conversely, the update area undergoes multiple noise and denoise steps for content generation.

\begin{equation}
\hat{z}_t = \hat{z}_t \odot   \mathcal{M} + z_t \odot (1 - \mathcal{M} ) 
\end{equation}

For consistency, rendered images from both the forward and backward views be aligned with the update area. When generating images with Depth2Image from two viewpoints, it's crucial to include the viewing angle information "forward viewing angle and backward viewing angle." This additional perspective detail helps reduce ambiguities in the Depth2Image model.

\subsubsection{Gradient descent optimizes Texture}
Firstly, We initialize a white texture map and rendering it from different viewpoints. The rendered image $I_r$ and the SD model-generated image $I_g$ are used to compute the loss for optimizing the texture map's gradients. For each viewpoint, we project the pixel coordinates of $I_r$ onto the texture map using UV mapping, followed by linear interpolation to determine the rendered pixel values. The texture map is then iteratively updated based on the MSE loss between $I_r$ and $I_g$. Throughout this process, UV mapping remains constant while the texture map is incrementally refined.

Gradient descent in differentiable rendering tends to introduce noise into texture maps, particularly noticeable in high-resolution outputs. This issue stems from the UV mapping process, where increased texture map resolution leads to greater dispersion of neighboring pixels' mapping coordinates. This dispersion, when exceeding the linear interpolation kernel's size, creates "point gaps" — unoptimized regions on the texture map. While these gaps don't affect rendering from the same viewpoint, due to consistent UV mapping with unchanged camera parameters, they become problematic during incremental updates for new views. These updates can lead to rendered pixels mapping into "point gaps", complicating gradient calculation and inducing noise. Our solution is a dual-optimization approach: alongside each incremental update, we refine a low-resolution texture map $t$ for self-supervised guidance and a high-resolution texture map $T$, mitigating the noise issue effectively.

\begin{equation}
    \mathrm{Loss} _{\mathrm{UV} } = \mathbf{MSE}\left (M_V\odot T',  M_V\odot t \right )  
\end{equation}

In the incremental update of the texture map, we initialize a resolution-independent triangular mesh texture mask $M_V$ and optimize a low-resolution texture map $t$ concurrently with the synthesis of the high-resolution texture $T$. The low-resolution map $t$ guides the global gradient for $T$. We downsample $T$ to $T'$, matching the size of $t$. Using $M_V$, the MSE loss between $T'$ and $t$ is computed, providing average gradient guidance to mitigate the discrete noise in $T$. This global optimization process updates the "point gap" regions with the average gradient, reducing the occurrence of "point gaps" and preserving the gradient backpropagation in the original mapping region.

\section{Experiments}

\begin{figure}[t]
  \centering
  \includegraphics[width=\linewidth]{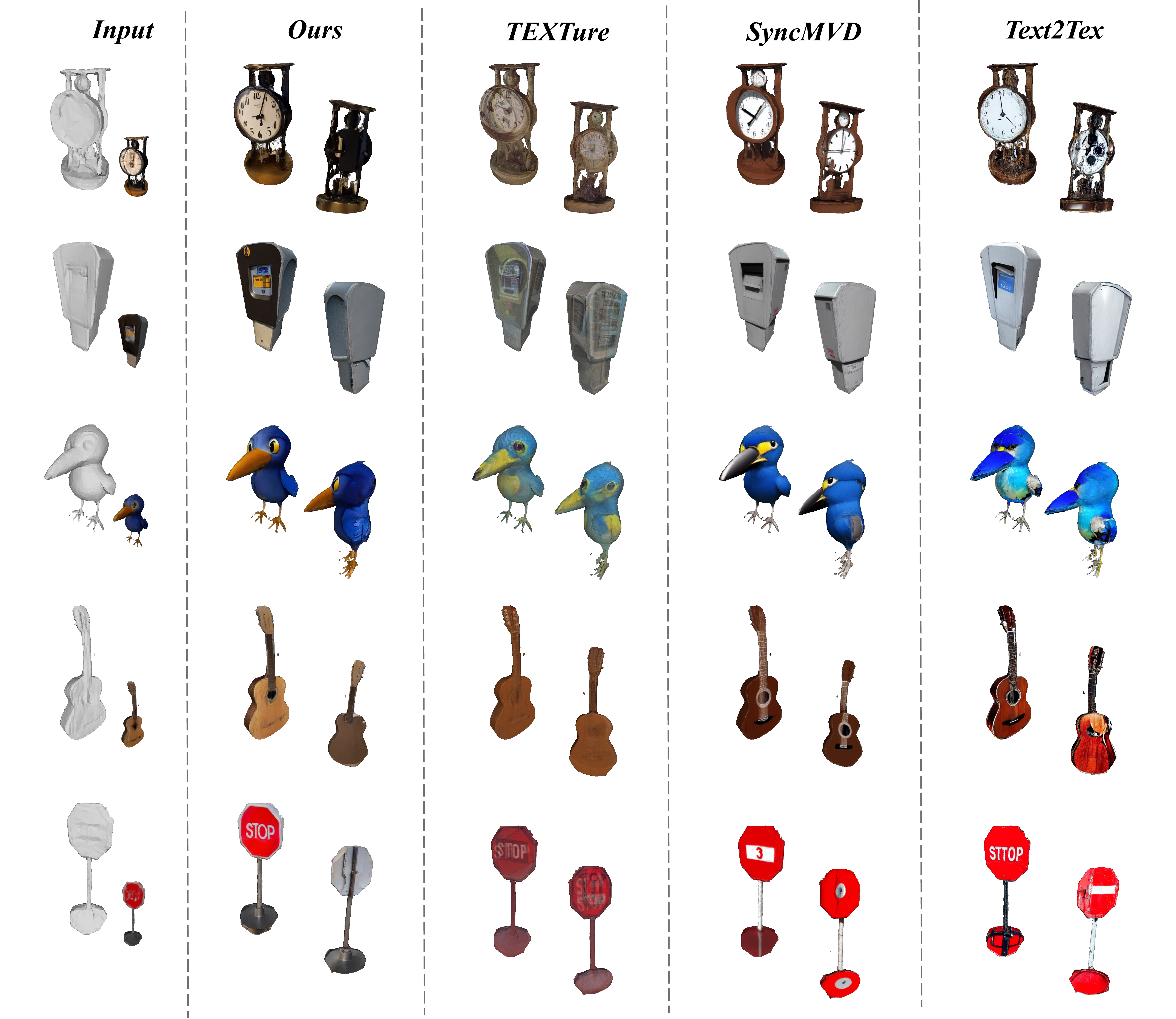}
  \caption{We demonstrate the comparison of texture synthesis effects under high-resolution conditions (4K). The textured mesh, obtained after structural simplification in the first stage, serves as the initial input for our method, whereas other methods start from a mere mesh template. Owing to this initial input, our approach exhibits fewer multi-face Janus problem and achieves a rendering effect with greater similarity to the initial texture.}
  \label{fig:result3}
\end{figure}

\subsection{Implementation Details}
In the texture optimization phase, we utilize ten camera views, encompassing the entire surround view along with top and bottom perspectives. The resolution for all rendered images is set uniformly at 1024x1024 pixels. For evaluating the final texture maps, we employ higher resolutions, specifically 1K and 2K. To improve the model's capacity for generating direction-specific views, we incorporate additional view descriptors such as "front" and "left front" when synthesizing with the SD model. Users have the flexibility to adjust the number of denoising steps in the SD model; however, for testing, we standardized the number of denoising iterations to 100 and set the number of iterative steps for gradient descent optimization of the texture map at 200. For meshes lacking UV mapping, we utilize Xatlas for UV calculations, which subsequently influences the gradient optimization process. Differentiable rendering is executed using the Kaolin library. The entire model is trained and tested on NVIDIA A100 GPUs.

\subsection{Quality Comparison Results}
In this section, we primarily discuss the comparison between the quality of our reconstructions and texture restorations. While reconstruction itself is not the central focus of our research, we place a significant emphasis on presenting comparative results related to high-definition (HD) texture restoration.

\subsubsection{Reconstruction quality comparison}
\begin{figure}[tb]
  \centering
  \includegraphics[width=0.8\linewidth]{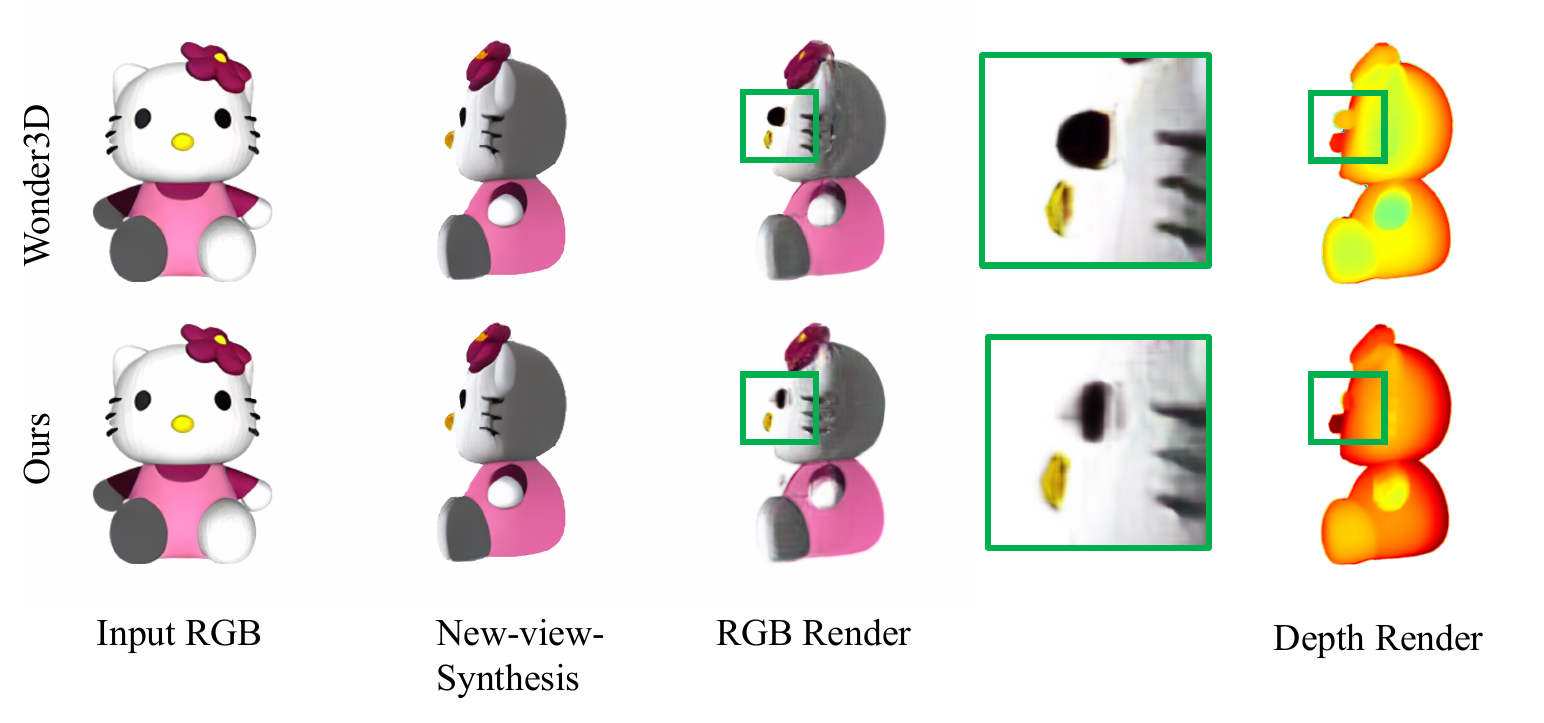}
  \caption{We present the rendering results of Wonder3D with and without depth supervision. Thanks to depth supervision, our method suppresses incorrect protrusions.}
  \label{fig:result2}
\end{figure}

We showcase our partial single-view reconstruction outcomes in Fig. \ref{fig:result1}. It is observed that the original Wonder3D framework encounters reconstruction inaccuracies on certain concave or protruding surfaces. To mitigate this, we have integrated depth supervision into the model. Unlike the local supervision provided by Wonder3D's normal supervision, which enhances surface detail, depth supervision offers a form of global oversight akin to color supervision. This global perspective aids in the overall shape and spatial disparity control, thereby enhancing the model's robustness. Our depth information is derived from the Depth anything model. Consistent with Wonder3D, we generate normals and viewpoint-consistent RGB images. The results demonstrate that depth supervision effectively reduces incorrect surface protrusions, as illustrated in Fig. \ref{fig:result2}.

\subsubsection{Texture recovery quality contrast}
The principal contribution of our scheme is the high-quality texture restoration it achieves. In comparison with conventional texture preservation algorithms and those like TEXTure \cite{richardson2023texture}, Text2Tex \cite{chen2023text2tex}, and SyncMVD \cite{liu2023text}, our method exhibits superior quality. We utilize the textured mesh obtained in the initial phase as the geometric input, then proceed to simplify this mesh's structure by reducing its face count to 3,000. This reduction simulates a typical user operation.

\begin{figure}[tb]
  \centering
\includegraphics[width=\linewidth]{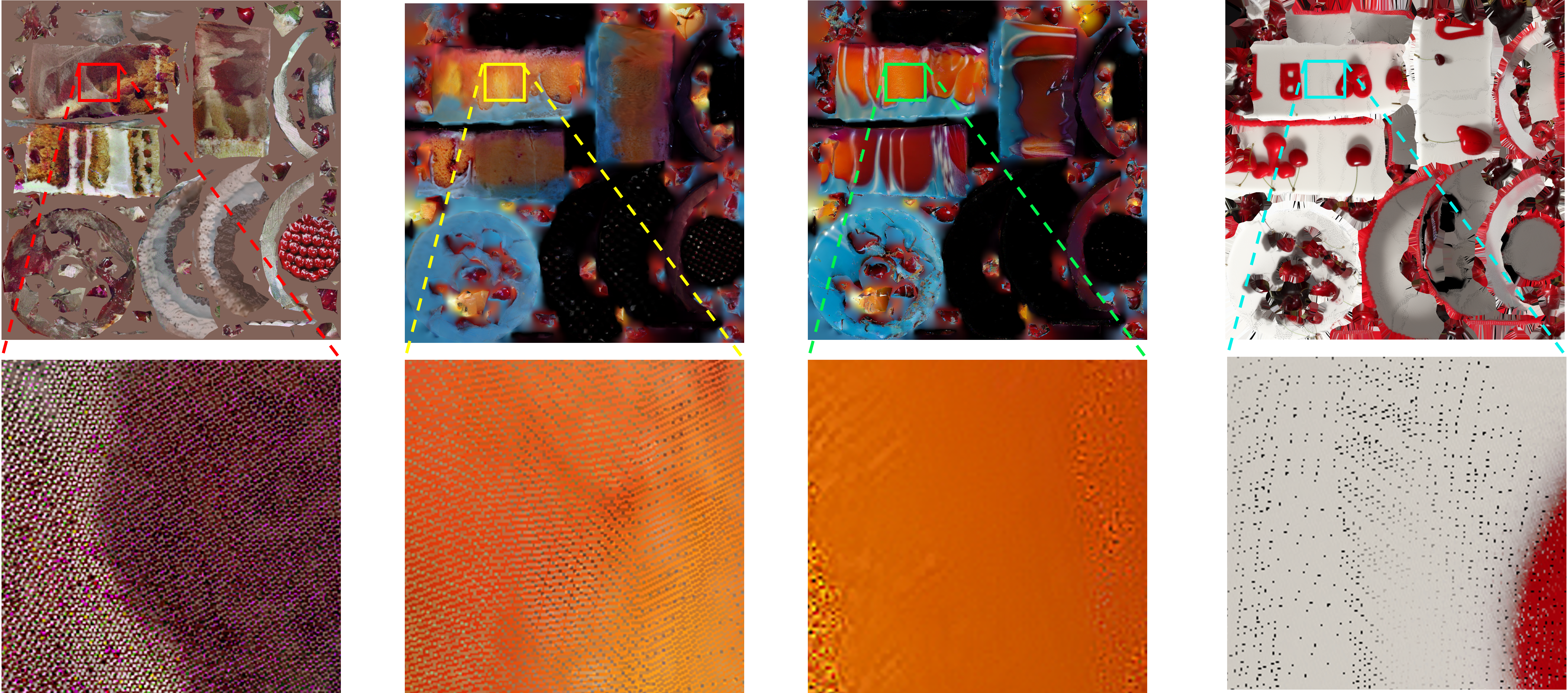}
  \caption{We show the TEXTure synthesis results in high resolution, from left to right: a)TEXTure texture synthesis. b) Texture texture synthesis with initial input. c) our method. d)SyncMVD texture synthesis results.}
  \label{fig:result4}
\end{figure}

In high-resolution (4K) texture synthesis, original methods often generate much noise, as shown in Fig. \ref{fig:result4}. Our approach employs a self-supervised technique that optimizes both low- and high-resolution texture maps concurrently, leveraging the low-resolution texture to refine the high-resolution counterpart. While SyncMVD \cite{liu2023text} mitigates texture noise by merging texture maps from various viewpoints, it does not entirely eliminate the issue. Conversely, Text2Tex \cite{chen2023text2tex} utilizes a viewpoint adaptive optimization strategy that iteratively adjusts the texture map from different angles to lessen noise, but this process is notably time-consuming, often extending beyond 15 minutes. Our method, in contrast, significantly reduces time consumption, completing within approximately 4 minutes. Moreover, by using user-adjusted rough textures as the initial input, our method greatly enhances the resemblance between the synthesized and original textures, akin to a high-fidelity restoration. The comparative effectiveness of our approach versus other methods is illustrated in Fig. \ref{fig:result3}.

\begin{figure}[tb]
  \centering
  \includegraphics[width=\linewidth]{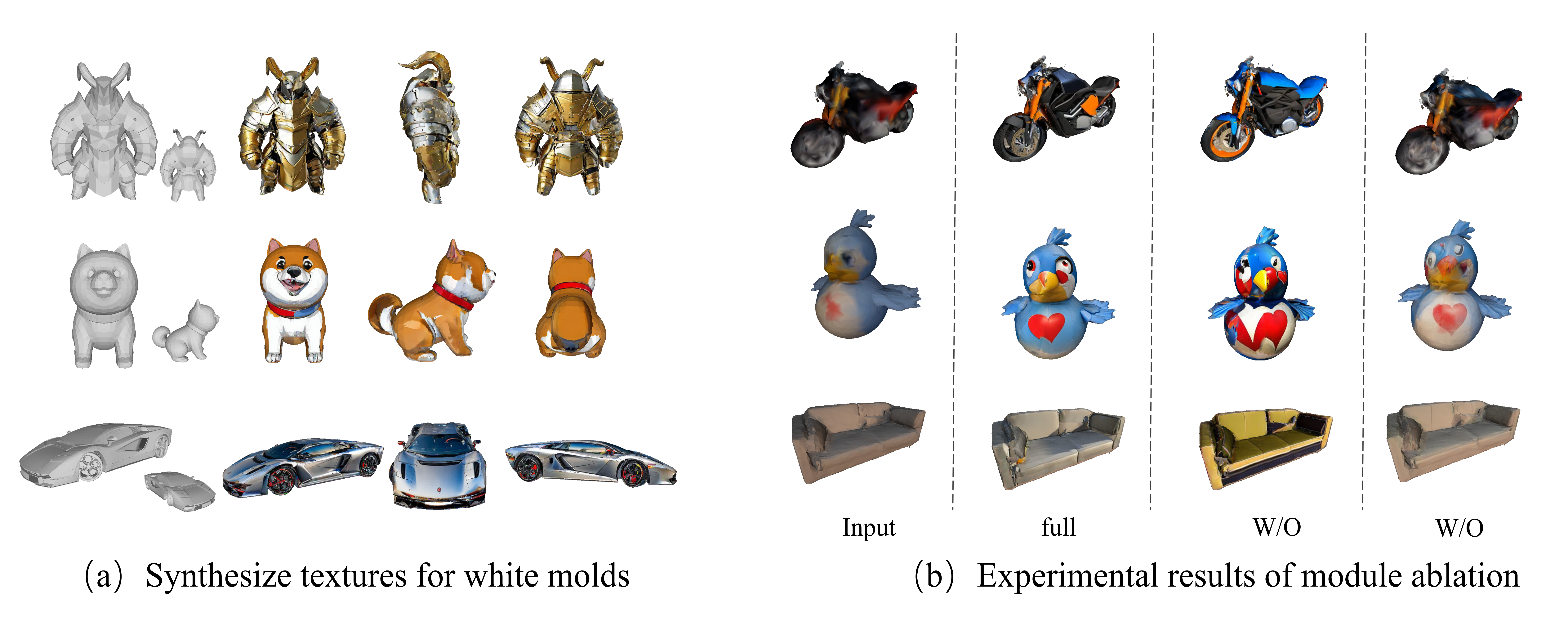}
  \caption{Visual Effects of Ablation Experiment. a) Demonstrates the ability of our scheme to synthesize textures for white models, i.e., removing the initial input. b) From left to right: initial input, complete scheme result, removal of checkerboard mask, removal of high and low-resolution self-supervision.}
  \label{fig:xiaorong}
\end{figure}

\subsection{Ablation Studies}
\subsubsection{Remove the initialization input}
Our approach utilizes a rough texture map, as manipulated by the user, as the initial input to facilitate the generation of a high-fidelity texture map that closely resembles the user's input. Should this initial texture input be omitted, our method essentially becomes a procedure for creating textures for untextured (white) models. To illustrate the efficacy of our method in applying textures to these white models, we conduct ablation studies using the geometry generated during the initial phase of our scheme. The outcomes of our texture synthesis process are displayed in Fig. \ref{fig:xiaorong}(a).

\subsubsection{Remove the Checkerboard Mask}
Our scheme utilizes the rough texture map after user action as the initial input to generate a high-fidelity texture map similar to the user's input. If this texture initialization input is removed, the entire scheme degenerates into a texture synthesis scheme for white models. We conduct ablation experiments with meshes downloaded from the Objaverse dataset to demonstrate our scheme's ability to apply textures to white models. We present our texture composition renderings in Fig. \ref{fig:xiaorong}(b).

\subsubsection{Remove self-supervision}
In our proposed scheme, high-resolution and low-resolution texture maps are synthesized simultaneously, with the primary function of the low-resolution texture maps being to eliminate noise generated during the synthesis of high-resolution texture maps. To verify the effectiveness of this self-supervised module, we synthesized only high-resolution texture maps, which inevitably generated noise and resulted in an overall fuzzy rendering effect. We illustrate the fuzzy rendering caused by this noise in.Fig. \ref{fig:xiaorong}(b).



\section{Limitations and Conclusions}
\subsection{Limitations}
Although our method successfully recovers high-fidelity textures, we have identified unresolved challenges. We observed errors in certain perspectives arising from the independent generation of images from varied viewpoints and subsequent reverse mapping, especially since the SD model tends to favor forward perspectives. We attribute these discrepancies primarily to the multi-view generation shortcomings of the SD model, which could be ameliorated by direct modifications to the UNet's attention mechanism. Additionally, as our texture maps are updated incrementally from different perspectives, seams are inevitably present, diminishing the quality of the resulting texture map. We believe this issue stems from the absence of global texture area coordination. Addressing these two concerns will be the focus of our forthcoming research and updates.

\subsection{Conclusions}
In this paper, we introduce a method for synthesizing high-fidelity textures at high resolutions from rough initial textures, offering a novel approach to texture restoration following model simplification. We employ a self-supervised approach to mitigate the noise issues inherent in existing high-resolution texture synthesis methods. Moreover, the simplified model's rough texture serves as the initial input for the Depth2Img model, thereby increasing the resemblance between the synthesized texture and the original texture. Essentially, the concept presented in this paper is broadly applicable to addressing noise issues in high-resolution texture synthesis and achieving high-fidelity outputs. The visual results demonstrate the efficacy of our method without significantly increasing time consumption. We aspire that our contribution will inspire future texture synthesis research and help overcome the current methods' constraints.

\bibliographystyle{splncs04}
\bibliography{ref}

\end{document}